\documentclass[letterpaper, 10 pt, conference]{ieeeconf} 
\IEEEoverridecommandlockouts  
\let\labelindent\relax

\usepackage{times,multirow,caption,float}
\usepackage{enumitem}
\usepackage{wrapfig}
\usepackage{graphicx}
\usepackage{epsfig,xspace,layout}
\usepackage{subfigure}
\usepackage{color}
\usepackage{amsfonts}
\usepackage{times}
\usepackage{amssymb}
\usepackage{amsmath}
\usepackage{rotating}
\usepackage{amsthm}
\usepackage{graphicx}
\usepackage{epsfig}
\usepackage{mathrsfs}
\usepackage{caption}
\usepackage{subfigure}
\usepackage{mathtools}
\usepackage{makeidx}
\usepackage{multirow} 
\usepackage{dblfloatfix}
\usepackage{threeparttable}
\usepackage{dsfont}
\usepackage[font={small}]{caption}
\usepackage{cleveref}
\usepackage{algorithm}
\usepackage{algorithmic}
\usepackage{tikz}
\usetikzlibrary{shapes,arrows,positioning,calc}
\usepackage{siunitx}
\usepackage{url}
\usepackage{cite}

\theoremstyle{definition}

\theoremstyle{remark}
\newtheorem{remark}{Remark}

\DeclareMathAlphabet{\mathpzc}{OT1}{pzc}{m}{it}

\DeclareFontFamily{U}{jkpmia}{}
\DeclareFontShape{U}{jkpmia}{m}{it}{<->s*jkpmia}{}
\DeclareFontShape{U}{jkpmia}{bx}{it}{<->s*jkpbmia}{}
\DeclareMathAlphabet{\mathfrak}{U}{jkpmia}{m}{it}

\DeclareMathOperator{\dt}{dt}

\newcommand{\R}{\mathbb{R}}

\renewcommand{\S}{\mathcal{S}}

\newcommand{\al}[1]{\begin{align}#1\end{align}}
\newcommand{\eq}[1]{\begin{equation}#1\end{equation}}
\newcommand{\ald}[1]{\begin{aligned}#1\end{aligned}}

\newcommand{\eqn}[1]{\begin{equation*}#1\end{equation*}}

\newcommand{\subeq}[1]{\begin{subequations}#1\end{subequations}}

\IEEEoverridecommandlockouts

\begin{document}
\title{Force-feedback based Whole-body Stabilizer for Position-Controlled Humanoid Robots}
\author{Shunpeng Yang$^{*}$, Hua Chen$^{*}$, Zhen Fu, and Wei Zhang\thanks{The authors are with the Department of Mechanical and Energy Engineering, Southern University of Science and Technology, Shenzhen, China.}
\thanks{$^*$These authors contributed equally to this work.}
\thanks{This work was supported in part by National Natural Science Foundation of China under Grant No. 62073159 and Grant No. 62003155, in part by the Shenzhen Science and Technology Program under Grant No. JCYJ20200109141601708, and in part by the Science, Technology and Innovation Commission of Shenzhen Municipality under grant no. ZDSYS20200811143601004.}
}
\maketitle

\begin{abstract}
This paper studies stabilizer design for position-controlled humanoid robots. Stabilizers are an essential part for position-controlled humanoids, whose primary objective is to adjust the control input sent to the robot to assist the tracking controller to better follow the planned reference trajectory. To achieve this goal, this paper develops a novel force-feedback based whole-body stabilizer that fully exploits the six-dimensional force measurement information and the whole-body dynamics to improve tracking performance. Relying on rigorous analysis of whole-body dynamics of position-controlled humanoids under unknown contact, the developed stabilizer leverages quadratic-programming based technique that allows cooperative consideration of both the center-of-mass tracking and contact force tracking. The effectiveness of the proposed stabilizer is demonstrated on the UBTECH Walker robot in the MuJoCo simulator. Simulation validations show a significant improvement in various scenarios as compared to commonly adopted stabilizers based on the zero-moment-point feedback and the linear inverted pendulum model.

\end{abstract}

\section{Introduction}
Legged robots have great potentials in numerous practical scenarios thanks to their outstanding ability in handling complex terrains. Among various categories of legged robots, humanoids have received a considerable amount of research attentions over the past several decades~\cite{kuindersma2016optimization, chen2020underactuated, castillo2020hybrid, 8630006,8814833, 6906613, chen2019optimal}. Depending on the types of input commands received by the joint actuators, humanoids are typically classified as position-controlled and torque-controlled. Generally speaking, position-controlled humanoids rely on harmonic gear based actuators to provide high torque output at relatively low response rates, which have been popular in both academia and industry over the past several decades. Examples include Honda ASIMO~\cite{hirose2007honda}, UBTECH Walker~\cite{WalkerWeb}, and AIST HRP-4~\cite{kaneko2008humanoid}. On the other hand, torque-controlled humanoids, such as Boston Dynamics Atlas~\cite{Atlas}, Agility Robotics Digit~\cite{Digit} and PAL Robotics TALOS~\cite{stasse2017talos} have become increasingly popular in recent years. They manage to achieve fast actuator response with sufficient torque output via new designs and actuation techniques.

Control of humanoid robots, especially position-controlled humanoids, is a very challenging task, mainly due to the complex whole-body dynamics, the underactuated floating-base, the complicated contact behavior, and the nonlinear and relatively slow joint servo responses. Currently, the control architecture for position-controlled humanoids mostly follows a hierarchical structure that involves motion planning, tracking control, and stabilizer. In the hierarchical framework, the motion planner is concerned with generating reference trajectories for the floating base and/or swinging foot that are compatible with a simplified template model (e.g., linear inverted pendulum model~\cite{973365} or spring-loaded inverted pendulum model~\cite{wensing2013high}) characterizing the humanoids' essential dynamics~\cite{Full3325}. Tracking controller, on the other hand, aims to determine input commands of joint actuators based on a more accurate model (e.g. whole-body model~\cite{wensing2014optimization}) to ensure that the robot's actual state closely follows the planned trajectory. Stabilizer, as a common module in the control architecture of position-controlled humanoids, mainly focuses on exploiting measurements of contact forces and robot states to design feedback schemes to further improve tracking performance~\cite{kajita2014introduction}. It has been shown that the stabilizer is crucial for reliable locomotion of position-controlled humanoids in both simulations and real-world experiments~\cite{5651082, 8794348}. 




As one of the pioneering works in stabilizer design, Hirai \emph{et al.} develops a stabilizer with ground reaction force control and ZMP control, where the former modifies the desired position and posture of the feet for driving the "'Center of Actual Total Ground Reaction Force" (C-ATGRF) to the desired zero-moment point (ZMP) and the latter changes the ideal body trajectory to shift the desired ZMP to an appropriate position~\cite{677288}. Sugihara and Nakamura~\cite{1041658} investigate the possibility of using center-of-mass (CoM) and ZMP feedback as well as vertical force feedback to stabilize CoM trajectory with CoM Jacobian. Kajita \emph{et al.} designs a stabilizer for the HRP-4C robot through directly modifying body posture according to vertical force distribution and modifying desired foot trajectories~\cite{5651082}, in which the linear inverted pendulum model is employed in the design. 
More recently, this strategy is augmented with divergent component of motion (DCM) feedback and is applied to achieve stair climbing with HRP-4 robot~\cite{8794348}. As for the stabilizer design of humanoid COMAN~\cite{tsagarakis2013compliant}, a passivity based admittance control scheme is formulated based on the cart-table model~\cite{1241826}. With such an approach, the CoM reference behaves as a spring-damper system with active compliance~\cite{zhou2014passivity}. Moreover, angular momentum modulation is also incorporated into the stabilizer design with a single rigid body model in~\cite{li2012passivity}.


Despite the rich literature, most existing stabilizer design strategies employ simplified models and simple ZMP measurement. These simplifications help streamlining the stabilizer design procedures, but are naturally limited in performance especially in challenging scenarios. Fundamentally speaking, effective stabilizers should make full use of the six-dimensional contact force measurement and coordinate the whole-body motion to achieve accurate tracking of the planned trajectory. To this end, this paper develops a novel force-feedback stabilizer based on a formal analysis of the whole-body dynamics of position-controlled humanoids subject to unknown contacts. 

The main contributions of this paper are summarized below. First and foremost, as compared to existing schemes that mainly rely on simplified models as well as ZMP feedbacks, the proposed stabilizer accounts for the whole-body dynamics and fully exploits the six-dimensional force measurement information. Second, this paper conducts rigorous analysis on the whole-body dynamics for position-controlled humanoids with unknown contact. The analysis result subsequently establishes the relationship between joint position command and the whole-body dynamics as well as the resulting contact force. Such a relationship serves as the fundamental guideline for our design of force-feedback based whole-body stabilizer. Third, by utilizing a kinematics counterpart of the above mentioned relation between the whole-body dynamics and the 6D contact force measurement, an optimization-based formulation of the proposed stabilizer is devised. The stabilizer requires solving two quadratic programs that can be efficiently implemented on real robots.

\section{Preliminaries and Overview of the Proposed Stabilizer Design}

\subsection{Dynamics of Position-Controlled Humanoid Robots}
Consider a generic humanoid robot whose equations of motion are governed by the multi-link rigid body dynamics:
\eq{
\label{eq: mcrb_dyn}
H(q)\ddot q + c(q,\dot q) = S^T\tau + J_c^T(q)F,
}
where $q = (q^{fb}, q^J) \in \R^{6+n_a}$ denotes the robot's configuration with $q^{fb} \in SE(3)$ representing the configuration of the floating base and $q^J\in \R^{n_a}$ representing the actuated joint configurations, $H(q) \in \R^{(6+n_a)\times (6+n_a)} $ is the generalized inertia matrix, $c(q,\dot{q})\in \R^{6+n_a}$ is the term addressing Coriolis, centripetal and gravitational effects, $\S\in \R^{n_a \times (6+n_a)}$ is the input selection matrix, $\tau \in \R^{n_a}$ denotes the torque inputs at the actuated joints, $J_c(q)\in \R^{6n_c \times (6+n_a)}$ is the contact Jacobian, and $F\in \R^{6n_c}$ collects all external contact forces with $n_c$ denoting the number of contacts. 

For position-controlled humanoid robots, the actuator torque $\tau$ cannot be directly regulated. Instead, they are determined by the joint servo controller $\phi$ denoted by 
\eq{
\label{eq:qJ_d_torque_map}
\tau = \phi(q^J_d,x_\phi).
}
The joint controller takes the commanded joint position $q_d^J$ as input and computes the joint torque $\tau$ based on the current joint position/velocity measurements and possibly other model information of the robot. In the above equation, $x_\phi$ accounts for the measurements and the controller internal states. For example, if a proportional–integral–derivative (PID) controller is used, then $x_\phi$ involves the current joint position $q^J$ and velocity $\dot q^J$ measurements, as well as the position error integral $\int (q_d^J - q^J)$. If a PD control with gravity compensation is used, then $x_\phi$ will also depend on $q^{fb}$ as the gravity is a function of $q$ in general. 

The exact form of $\phi$ in~\eqref{eq:qJ_d_torque_map} is often not available, and is irrelevant to our design. However, it allows us to see how the joint position command affect the system dynamics. With equation~\eqref{eq:qJ_d_torque_map}, the dynamics of position-controlled robots with position commands $q^J_d$ become:
\eq{
\label{eq: position-based wb dyn}
H(q)\ddot q + c(q,\dot q) = S^T\phi(q^J_d,x_\phi) + J_c^T(q)F.
}
\begin{figure*}[tp!]
    \centering
    \includegraphics[width=0.95\linewidth]{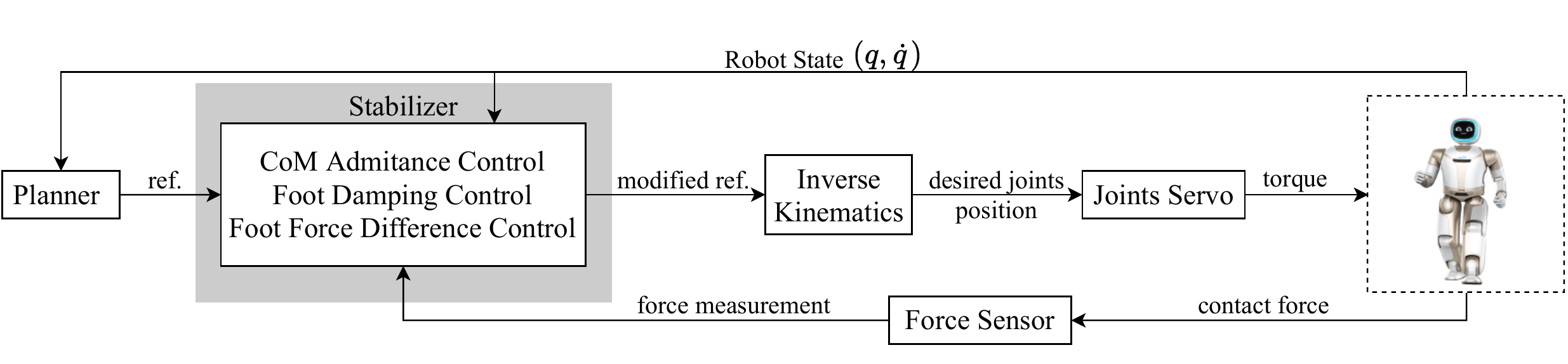}
    \caption{The flow diagram of the existing control framework}
    \label{fig: traditional-stabilizer-design}
    \vspace{-10pt}
\end{figure*}
\begin{figure*}[tp!]
    \centering
    \includegraphics[width=0.97\linewidth]{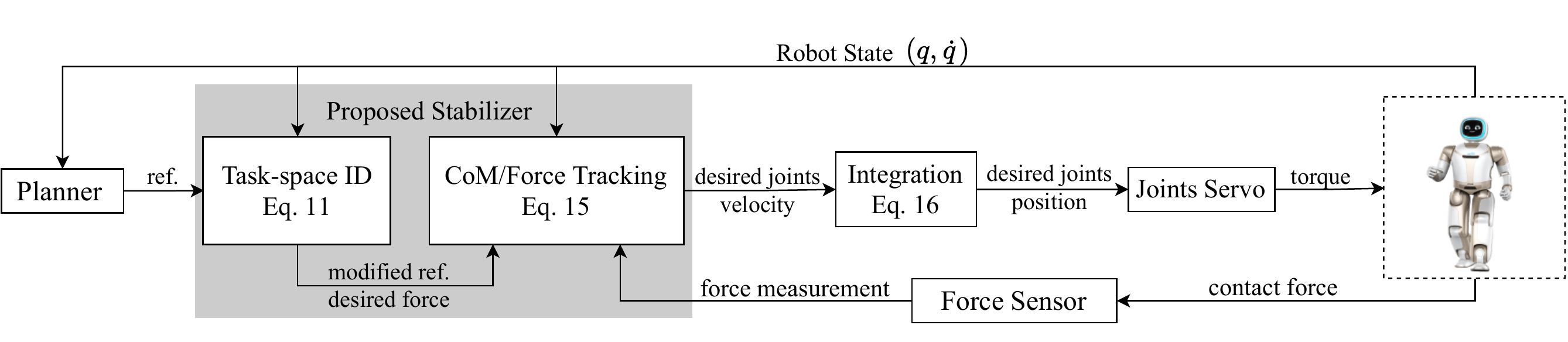}
    \caption{The complete control framework with the proposed stabilizer based on force feedback.}
    \label{fig: stabilizer-design-force-feedback}
    \vspace{-10pt}
\end{figure*}
Due to the inherent nonlinearities of the actuator model, along with various other issues such as the complicated contact and underactuated floating-base dynamics, it is in general challenging to design controllers for position-controlled humanoids based on this dynamic model.

\subsection{Preliminaries on Existing Control Schemes}
To account for the above mentioned challenges in designing controllers for position-controlled humanoids, hierarchical schemes (Fig.~\ref{fig: traditional-stabilizer-design}) are commonly adopted. In such schemes, simplified models such as linear inverted pendulum (LIP) and variable height inverted pendulum (VHIP) are commonly adopted to generate reference trajectories of the robot's CoM (floating base) and feet motions. Then, these reference trajectories are used as inputs to an inverse kinematics (IK) solver that determines the corresponding commanded joint angles for the actual robot. To account for uncertainties, reference trajectories generated by the motion planner need to be modified based on real-time robot state and contact force measurements. These are achieved with {\em stabilizers} which have been widely recognized as a critical module in achieving successful locomotion with position-controlled humanoids.

Albeit the great accomplishments of the existing stabilizers, there remains several critical issues that prevent existing stabilizers to excel in challenging scenarios. Lack of consideration of whole-body dynamics and inadequate usage of the six-dimensional contact force measurement are among the primary limitations. In addition, considering only simplified models and decoupling the stabilizer design for different tasks fail to account for the coupling effects that can be  utilized to further improve the overall performance. . Moreover, encoding the full six-dimensional contact force information with the two-dimensional ZMP position inevitably loses critical measurement information.  

To address these issues, we develop a novel force-feedback based whole-body stabilizer that fully exploits the contact force information and accommodates whole-body dynamics of the position-controlled humanoids. In what follows, we first give a brief overview of the proposed scheme.

\subsection{Overview of the Proposed Stabilizer Architecture}

The schematic diagram of the proposed control architecture, particularly the stabilizer is shown in Fig.~\ref{fig: stabilizer-design-force-feedback}.
When the high-level references are generated by motion planning, the trajectory is first filtered by an inverse dynamics solver in the proposed stabilizer, which outputs the reference contact force to be tracked by the actual robot. Then, the modified references and the associated contact force references are fed into a task-space differential inverse kinematics solver that generates the commanded joint velocity for tracking the planned trajectories. The commanded joint velocity subsequently yields the commanded joint position through a simple integration and then is finally transmitted to the joint servo to control the actual humanoid robot.

Different from most existing schemes, the proposed stabilizer jointly considers the CoM tracking and contact force tracking tasks, by incorporating them into one quadratic program (QP). In the sequel, we detail our development of the proposed force-feedback based whole-body stabilizer by first rigorously analyzing the whole-body dynamics of the position-controlled humanoid with unknown contacts and then devising a QP based scheme for implementing the desired stabilizer. 


\section{Analysis of Whole-body Dynamics with Unknown Contacts}

Principled design of effective stabilizers for position-controlled humanoid robot requires careful analysis and thorough understanding of how the CoM and contact force behave in response to input to the humanoid robot (position commands) within unknown environments. In particular, provided the input-to-output relationship, i.e., how the joint position commands would affect the robot's CoM behavior and the actual contact force, synthesis of reliable stabilizers can be easily achieved via optimization based techniques. 

Motivated by the above discussion, this section first investigates how the joint position commands affect the CoM behavior as well as the actual contact force at the dynamics level, then derives a kinematics counterpart of this dynamic relationship that preserves the central connection between joint commands and contact forces. Utilizing this kinematics level relationship, the stabilizer design problem is decomposed into a reference generation problem with an inverse dynamics formulation and a task-space differential inverse kinematics problem, where both problems can be efficiently solved as quadratic programs.

To initialize our discussion, note that dynamics of position-controlled humanoids with nominal contact are given by 
\subeq{\al{\label{eq:exp_dyn1} & H(q)\ddot q + c(q,\dot q) = S^T\phi(q^J_d,x_\phi) + J_c^T(q)F, \\ \label{eq:exp_dyn2} & \Phi(q) = \text{const.} }\label{eq:exp_dyn}}
where~\eqref{eq:exp_dyn1} is basically the whole-body dynamics with position-controlled joints, and~\eqref{eq:exp_dyn2} stems from the fixed contact point condition. Differentiating~\eqref{eq:exp_dyn2} twice results in the following dynamics constraint \eqn{ J_c(q)\ddot{q} +\dot{J}_c(q) \dot{q} = 0.}

In practice, especially when the terrain is potentially uneven and unknown to the robot, the above model fails to accurately characterize the robot's dynamics with contacts. To account for the unavailability of the exact terrain configuration, we relax the fixed contact point constraint and instead allow the contact point to depend on both time and the robot's configuration as follows
\eq{\label{eq:v_contact}\Phi(q) + \epsilon(q,t) = 0.} 

Analogous to the previous case, differentiating this modified contact point constraint twice yields the following contact point acceleration constraint for system dynamics
\eq{\label{eq:general_contact}\dot J_c(q)\ddot q + J_c(q)\dot q + \eta(q,\dot{q},\ddot{q},t) = 0,} where 
\eq{
\eta(q,\dot{q},\ddot{q},t) = \ddot \epsilon(q,t)\nonumber}
denotes the generic modification of the acceleration at the contact point. In the sequel, we drop the dependence of $\eta$ on all the arguments for simplicity.

\begin{remark}\label{rmk:csa}
The quantity $\eta$ in~\eqref{eq:general_contact} is closely related to the \emph{contact separation acceleration} by Featherstone~\cite{featherstone2014rigid}. In fact, the \emph{contact separation acceleration} mainly focuses on the linear acceleration along the normal direction to contact, while the adopted $\eta$ accounts for the six-dimensional (both position and orientation) acceleration of the contact point. 
\end{remark}

With this adjusted model related to contact interaction, the relationship between input joint position commands $q_d^J$ and the resulting contact force $F$ can be easily established. By simply solving $\ddot{q}$ from~\eqref{eq:exp_dyn1} and plugging the expression into~\eqref{eq:general_contact}, the following key relationship is obtained:
\eq{\label{eq:force_joint}F =\Gamma_c(q)\phi(q^J_d,x_\phi) - \Lambda_c(q)\eta +  h_c(q,\dot q),} where \eqn{ \Lambda_c(q) = \left( J_c(q)H^{-1}(q)J_c^T(q)\right)^{-1}} is the so-called \emph{contact inertia}, and $\Gamma_c(q)$ and $h_c(q,\dot{q})$ are then given by \eqn{\ald{\Gamma_c(q)  & = -\Lambda_c(q)J_c(q)H^{-1}(q)S^T \\  h_c(q,\dot q) &= \Lambda_c(q)\left(J_c(q)H^{-1}(q)c(q,\dot q) - \dot J_c(q)\dot q\right). }}
Given the above expression~\eqref{eq:force_joint}, if the exact forms of $\eta$ and $\phi$ are known, the desired relationship between the input joint position commands $q_d^J$ and the contact force $F$ can be obtained. However, it is in general impossible to acquire the exact form of $\eta$ that characterizes the real contact interaction.

Aiming at establishing direct relationship between the input joint position commands $q_d^J$ and the contact force $F$, we first need to eliminate the effect of $\eta$. To this end, note that under the expected contact scenario, i.e., ${\eta} = 0$, any reference contact force $F_\text{ref}$ and its associated reference joint command $q_\text{ref}^J$ should satisfy the following constraint
\eq{F_\text{ref} = \Gamma_c(q)\phi(q_\text{ref}^J,x_\phi) + h_c(q,\dot{q})\label{eq:ref_force} .}

Now, given any generic robot state $(q,\dot q)$, the relationships~\eqref{eq:force_joint} and~\eqref{eq:ref_force} essentially dictate the constraints on how the actual humanoid and how the reference should behave. By subtracting~\eqref{eq:ref_force} from~\eqref{eq:force_joint}, the force error $F-F_\text{ref}$, if available from measurements, can provide useful information about the actual contact scenario characterized by $\eta$ as follows
\eq{F-F_\text{ref} = \Gamma_c(q)(\phi(q^J_d)-\phi(q^J_\text{ref})) -\Lambda_c(q)\eta.}

\begin{remark}
The above relationship indicates that, the actual contact interaction can be accurately estimated by comparing the actual contact force measurement with a reference contact force compatible with the nominal contact interaction constraint, provided that the actuator servo model is known.
\end{remark}

Substituting the above expression into~\eqref{eq:general_contact} yields the constrained joint accelerations compatible with the actual contact force and any feasible reference pair $(F_\text{ref},q^J_\text{ref})$:
\eq{J_c\ddot q\! +\! J_c\dot q\! = \!-\Lambda_c^{-1}(F\! -\! F_\text{ref})\! +\! \Lambda_c^{-1}\Gamma_c\left(\phi(q^J_d)-\phi(q^J_\text{ref})\right) \label{eq:contact_measure}}

So far, through analyzing the whole-body dynamics of position-controlled humanoid robot and the associated contact constraints, we have unveiled how the input joint position commands affect the robot's dynamics and the contact force. Nevertheless, due to the lack of knowledge about the accurate joint actuator servo model $\phi$ and the remaining freedom of adjusting the reference contact force $F_\text{ref}$, the above relationship~\eqref{eq:contact_measure} cannot be directly applied in our stabilizer design. In the following Section, we first detail how the reference contact force $F_\text{ref}$ is selected to account for tracking of other tasks such as the CoM, and then discuss how~\eqref{eq:contact_measure} can be further simplified, which jointly lead to our proposed kinematics based stabilizer. 


\section{Task-space Differential Inverse Kinematics based Stabilizer Design}

The proposed stabilizer adopts a differential inverse kinematics (Differential IK) based scheme, which computes the joint velocities first and then determines the input joint position commands by integrating the joint velocities. Such a scheme, instead of directly working with the conventional nonlinear inverse kinematics, takes advantage of the linear relationship between the joint velocities and the end-effector (feet) velocities that in turn allows for a tractable quadratic programming based approach.

\subsection{Desired Contact Force Generation via Inverse Dynamics}
Recall that the primary objective of the stabilizer is to improve the performance of tracking the desired CoM $x_d$ and the desired contact force, where the desired CoM is provided by high-level motion planner while the desired contact force remains unspecified. Hence, the first step of our stabilizer design is to generate a desired contact force that respects the dynamic constraint~\eqref{eq:ref_force} and tracks the desired CoM.

The proposed stabilizer leverages the inverse dynamics based scheme to generate the desired contact force via solving the following quadratic programming:
\subeq{\label{op: tsid}\al{\min_{\ddot q, \tau, F}\quad &\sum_{i=1}^N\|J_i(q)\ddot q+\dot J_i(q)\dot q - \ddot y_d^i\|_{Q_i} + \|\tau\|_{Q_\tau}, \\
\text{s.t.} \quad &H(q)\ddot q + c(q,\dot q) = S^T\tau + J_c^T(q)F, \\ \label{eq:tsid_act_fric}
    & \dot J_c(q)\dot q + J_c(q)\ddot q = 0,\\
    & H_\tau \tau \le 0, H_F F \le 0,}}
where $y^1_d = x_d$ accounts for the CoM task in particular, and $y^i_d$ $(i=2,3,...,N)$ denotes the other task references such as foot swinging. In the above inverse dynamics problem, the constraint~\eqref{eq:tsid_act_fric} takes care of the joint actuation limit and friction cone restriction. The above problem is solved online to generate the reference contact force, denoted by $F_\text{ID}$, that will then be fed into the differential IK based tracking scheme. 

\subsection{CoM and Contact Force Tracking via Differntial IK}
To account for tracking of both the desired CoM and desired contact force, we essentially need to identify how the CoM and contact interactions are influenced by the commanded joint velocities. Owing to the underactuated nature of the floating base of the humanoid system, the CoM velocity can only be adjusted provided that desired contacts are achieved. Hence, our stabilizer design considers the CoM velocity itself as a decision variable that is constrained and optimized in compliance with the contact task. 

The core of the developed stabilizer is the successful incorporation of all tasks into a unified problem, especially the contact force task. By simply integrating the critical relationship~\eqref{eq:contact_measure} with the desired force $F_\text{ID}$ plugged in, we obtain the following kinematics level relationship between the contact force and velocities of the configuration variables. 
\eq{\label{eq:kine_cons_force} \ald{J_c\dot{q} = &-\Lambda_c^{-1}\int \left( F-F_\text{ID}\right) \dt + \\  & \qquad \qquad \int  \Lambda_c^{-1}\Gamma_c\left(\phi(q^J_d)-\phi(q^J_\text{ID})\right) \dt  }} where $q^J_\text{ID}$ is the joint position command associated with $F_\text{ID}$.

Due to the unavailability of the actuator servo model $\phi(\cdot)$, the last term in~\eqref{eq:kine_cons_force} is viewed as a disturbance term hereafter in our stabilizer design that will be further minimized in the optimization based differential IK stabilizer. 

Denoting by $e_F = F -F_\text{ID}$ the force error between actual contact force and the desired one, we look for finding the input to ensure successful tracking of the desired contact force that achieves the following closed-loop error dynamics
\eq{\label{eq:des_error_dyn} \dot{e}_F = K  e_F,}
where $K \in \R^{6n_c\times 6n_c}$ is a negative-definite gain matrix that ensures convergence of the desired error dynamics. Integrating the above desired closed-loop dynamics and then substituting the result into~\eqref{eq:kine_cons_force} yields the following relationship between force error $e_F$ and joint velocities $\dot q$
\eq{\label{eq:cons_kine} J_c \dot q = -\Lambda_c^{-1}  K^{-1}e_F + \Delta,} where $\Delta=  \int \Lambda_c^{-1}\Gamma_c\left(\phi(q^J_d)-\phi(q^J_\text{ID}))\right) \dt $. 
\begin{remark}\label{rmk:adc}
Equation \eqref{eq:cons_kine} exactly implies a admittance-control law. However, what differs from the conventional form is the feedback gain related to current robot configuration by the contact inertia matrix $\Lambda_c$, which could be regarded as a adaptive gain rather than a fixed one like in traditional form.
\end{remark}

Now, the stabilizer design problem amounts to the determination of $\dot q$ to ensure tracking of both the desired contact force $F_\text{ID}$ and other references such as CoM, provided with the real-time measurement of the robot's configuration $q$, and the contact force error $e_F$. Let $\dot q = (v^{fb},v^J)$ be the decision variable containing both the CoM velocity $v^{fb}$ and all joint velocities $v^J$, solving the following task-space differential inverse kinematics simultaneously achieves contact force tracking as well as fulfillment of other tasks
\subeq{\label{op: u}\al{
\min_{v^{fb},v^J,\Delta} \quad & \sum_{i=1}^N\|J_i(q)\begin{bmatrix}
v^{fb} \\ v^J\end{bmatrix} - \dot y_d^i\|_{Q_i} + \|\Delta\|_{Q_d} \\
\text{s.t.}  \quad & J_c(q)\begin{bmatrix}
 v^{fb} \\ v^J
\end{bmatrix} =-\Lambda_c^{-1}  K^{-1}e_F + \Delta \label{eq:dik-cons1} \\ &
\underline{v^J} \le v^J \le \overline{v^J}, }} 
where the first term in the cost function accounts for the tracking of the desired CoM reference and other tasks. The second term in the cost function jointly with the first constraint~\eqref{eq:dik-cons1} ensures that the optimized CoM and joint velocities spare their maximal effort in achieving a closed-loop contact force error dynamics as specified in~\eqref{eq:des_error_dyn}. 

Once the velocity commands are obtained from solving the above quadratic program, the input joint position commands at time $t$ are then determined by the following dynamics 
\subeq{\al{\dot{q}_d^J(t) &= v^J\\ q_d^J(0)& = q^J(0),}}
which completes the overall stabilizer design.

\section{Simulation Experiments}
\begin{figure*}[htbp!]
    \centering
    \includegraphics[width=0.9\linewidth]{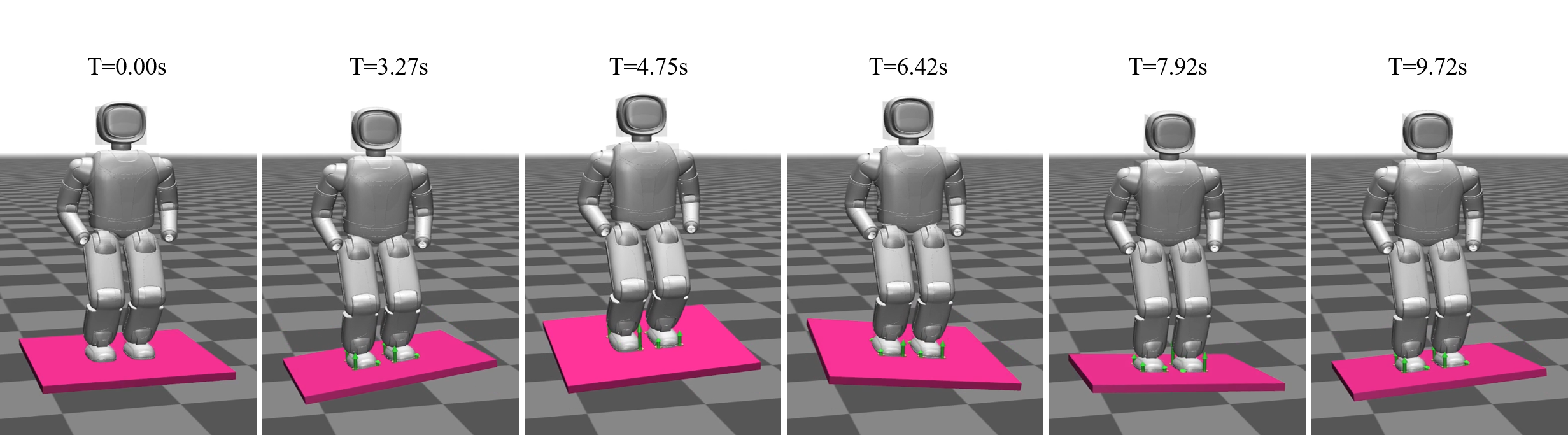}
    \caption{The height and attitude of the platform are changed at the same time.}
    \label{fig:heightAndRot}
\end{figure*}

\begin{figure*}[htbp!]
\centering
\subfigure[Position of floating base]{
    \includegraphics[width=0.225\linewidth]{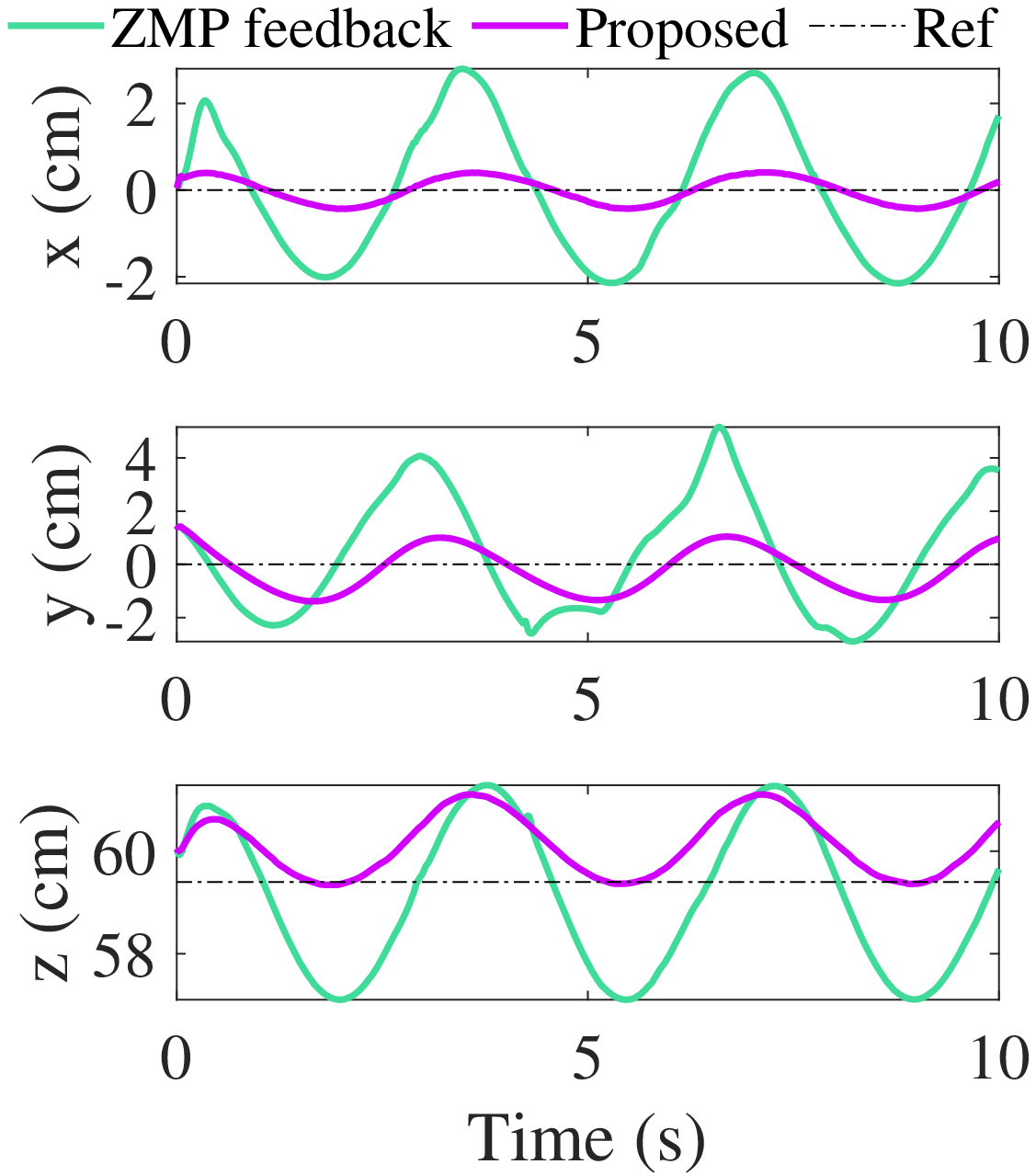}
    }
\subfigure[Rotation of floating base]{
    \includegraphics[width=0.225\linewidth]{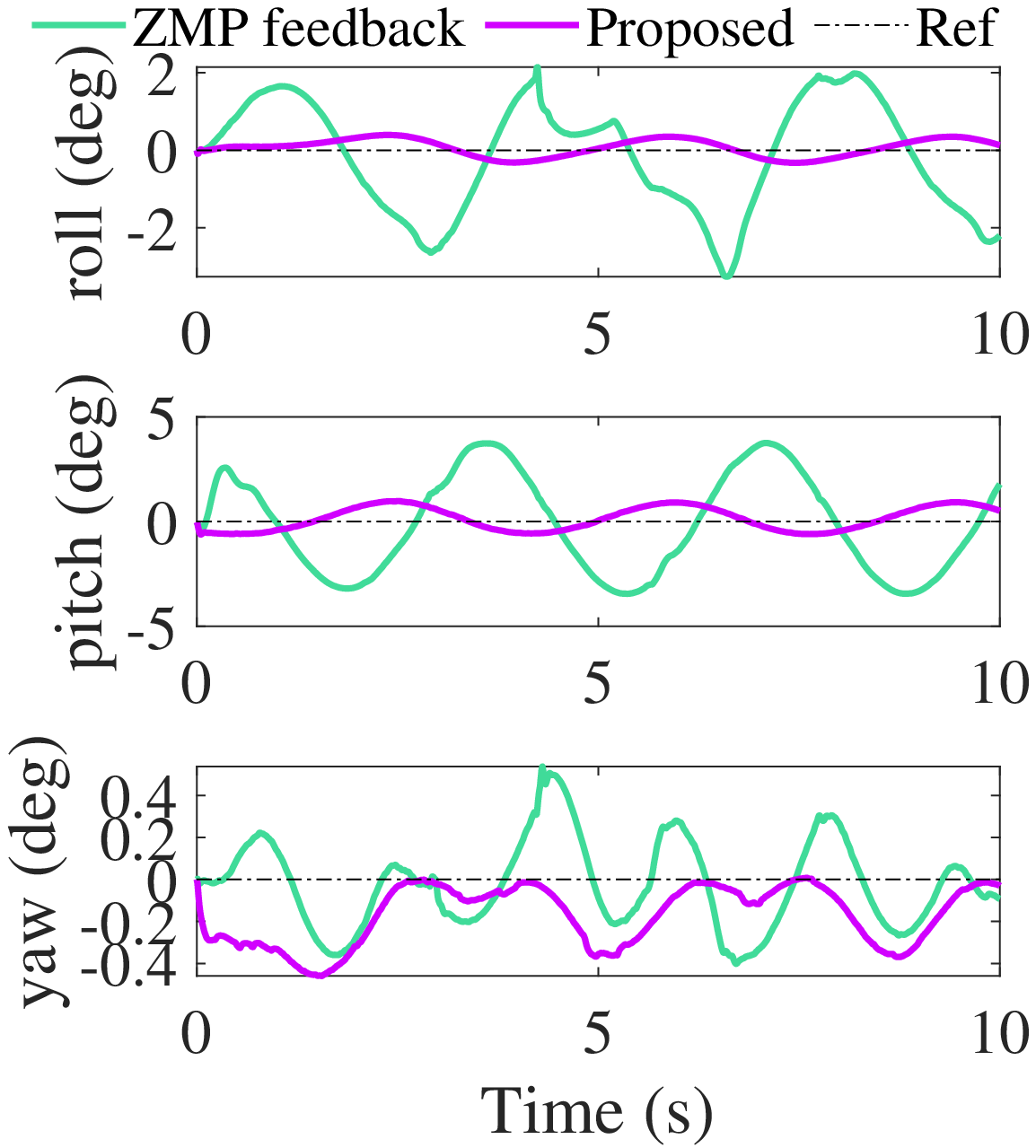}
    }
\subfigure[Force on the left ankle]{
    \includegraphics[width=0.245\linewidth]{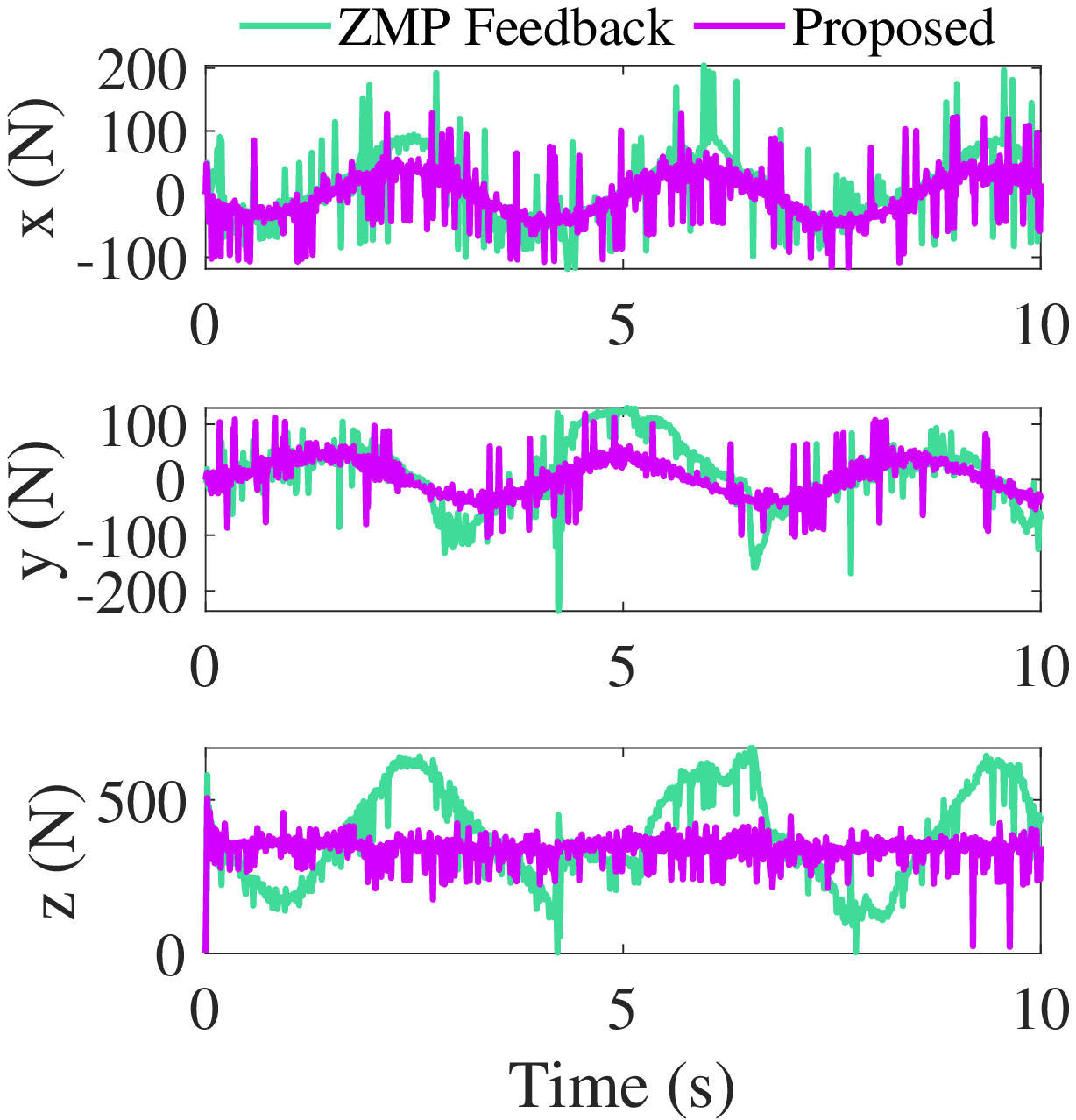}
    }
\subfigure[Moment on the left ankle]{
    \includegraphics[width=0.225\linewidth]{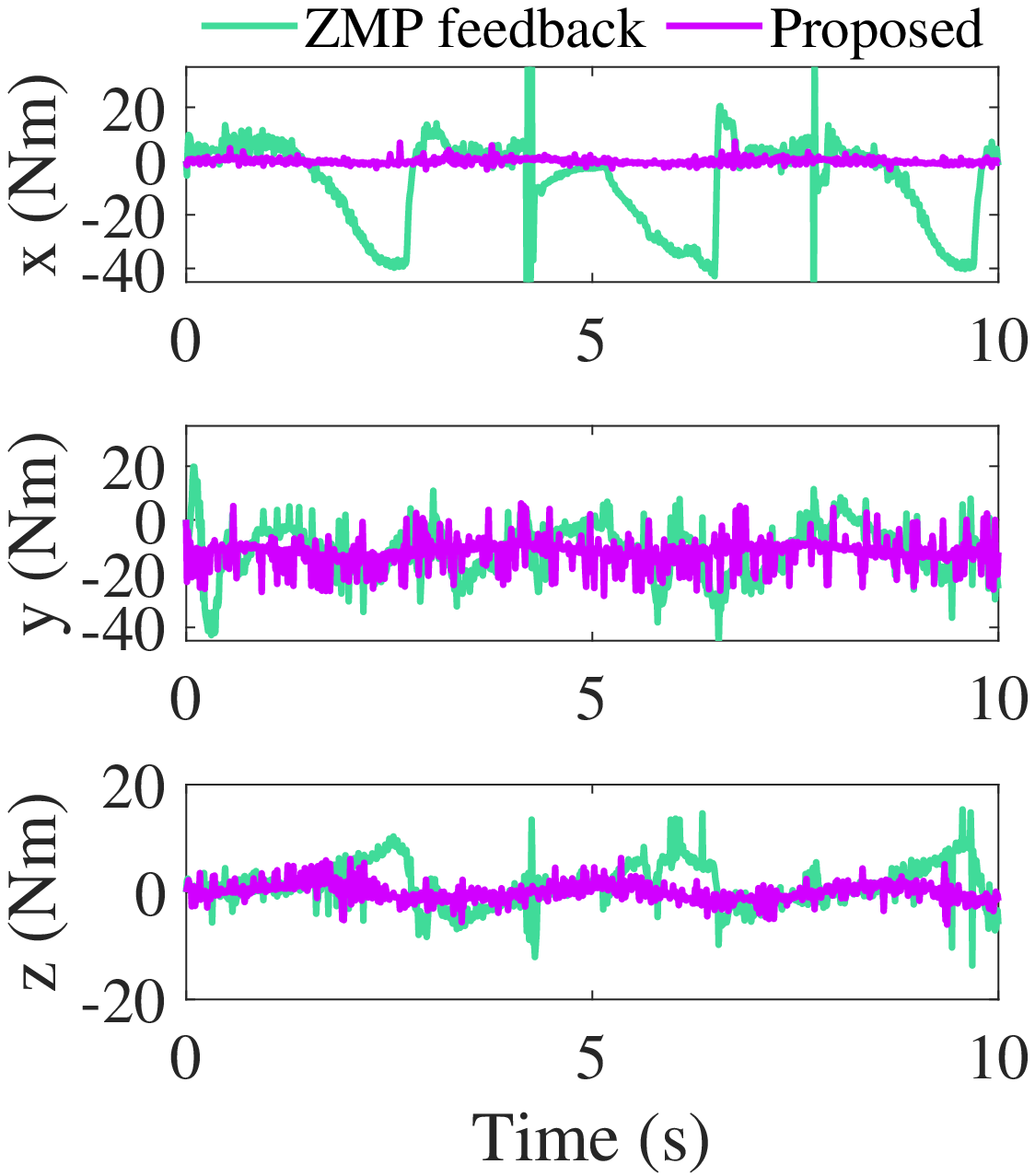}
    }
\caption{Floating-base trajectory and ankle wrench measurements corresponding to the scene shown in Fig.~\ref{fig:heightAndRot}}
\label{fig: hr_traj}
\end{figure*}
Performance of the novel force-feedback based whole-body stabilizer (\eqref{op: tsid} and~\eqref{op: u}) is validated using the UBTECH Walker robot in MuJoCo~\cite{todorov2012mujoco} with friction coefficient $\mu=0.7$. In addition, a discrete-time version of the proposed stabilizer is implemented, with time step $\delta t=0.001$s. The UBTECH Walker is a $\SI{1.5}{m}$ tall, $\SI{70}{kg}$ weight position-controlled humanoid with 28 actuated joints. Detailed parameters of Walker robot could be found on GitHub~\cite{Walker}. In all the simulation experiment, Pinocchio~\cite{pinocchioweb} is used to compute the physical quantities required in the proposed framework and the TSID library~\cite{del2016implementing} is used to compute the task-space inverse dynamics~\eqref{op: tsid}. For comparison, a ZMP-based stabilizer~\cite{8794348} is implemented as the baseline. 



\subsection{Scenario \uppercase\expandafter{\romannumeral1}: Balancing on a Moving Platform}

\begin{figure*}[htbp!]
    \centering
    \includegraphics[width=0.9\linewidth]{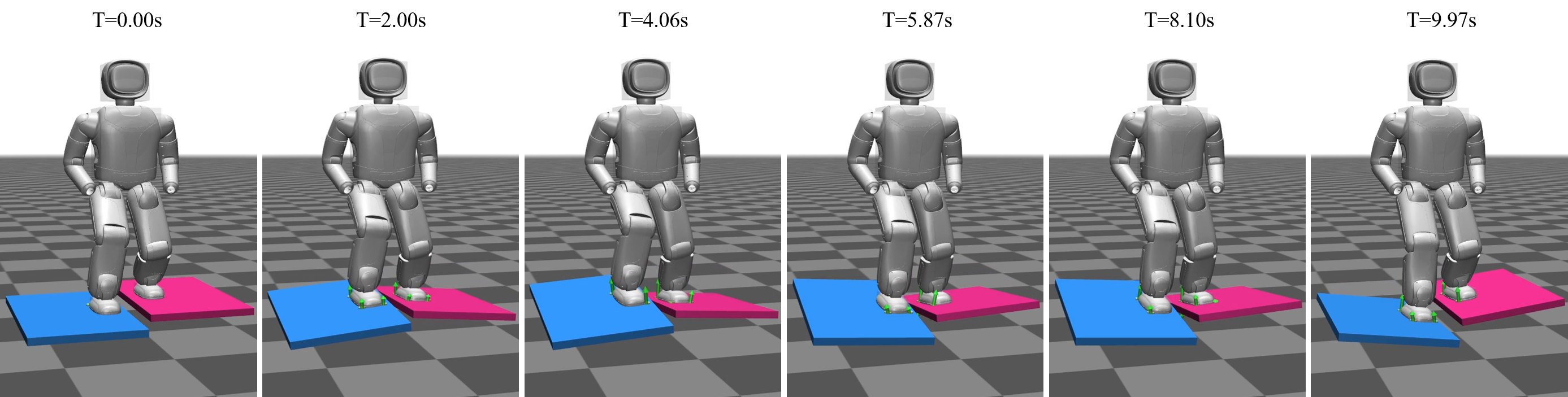}
    \caption{The height and attitude of the two platforms are changed separately at the same time.}
    \label{fig:double}
\end{figure*}

\begin{figure*}[htbp!]
\centering
\subfigure[Position of floating base]{
    \includegraphics[width=0.215\linewidth]{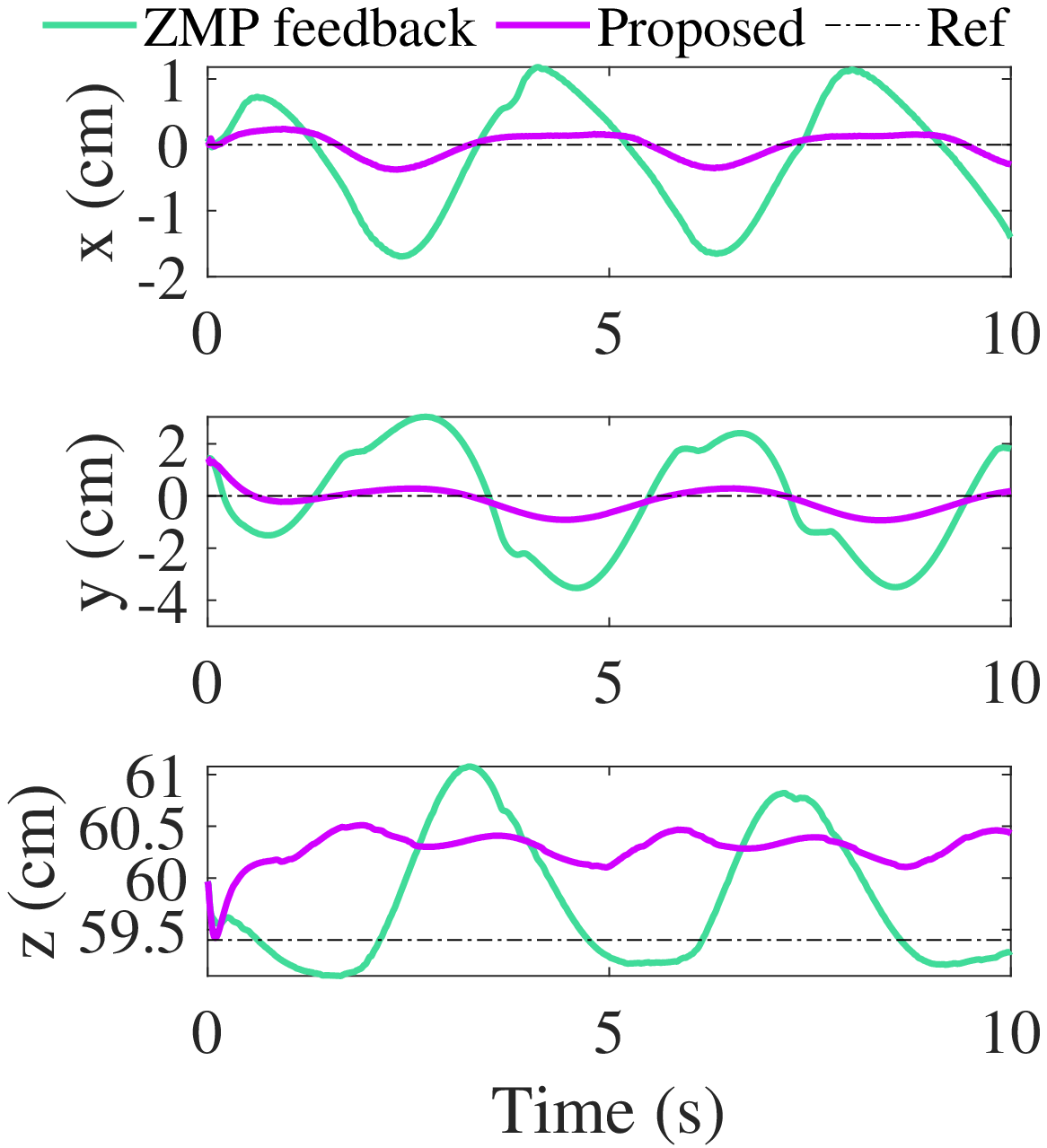}
    }
\subfigure[Rotation of floating base]{
    \includegraphics[width=0.215\linewidth]{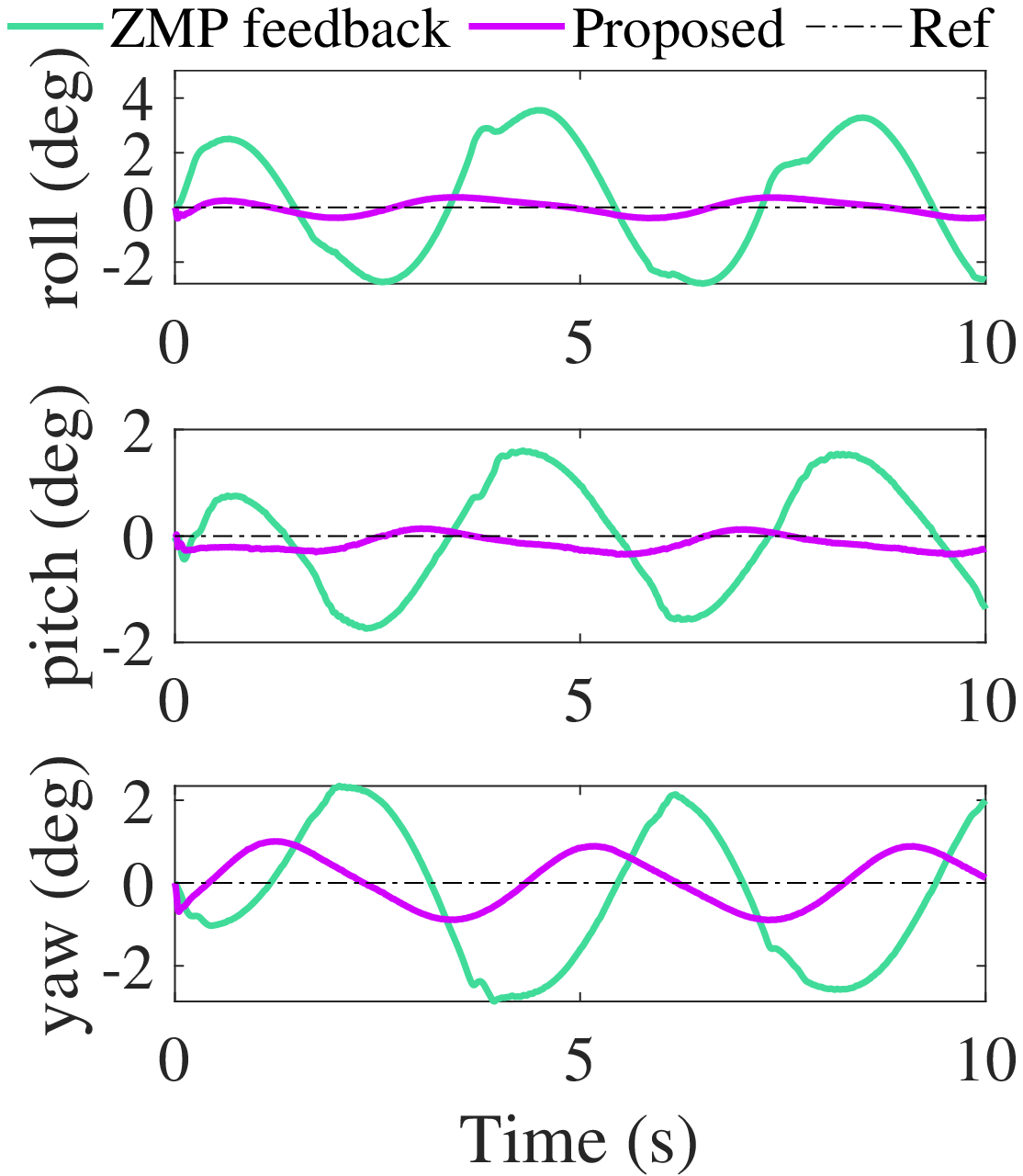}
    }
\subfigure[Force on the left ankle]{
    \includegraphics[width=0.235\linewidth]{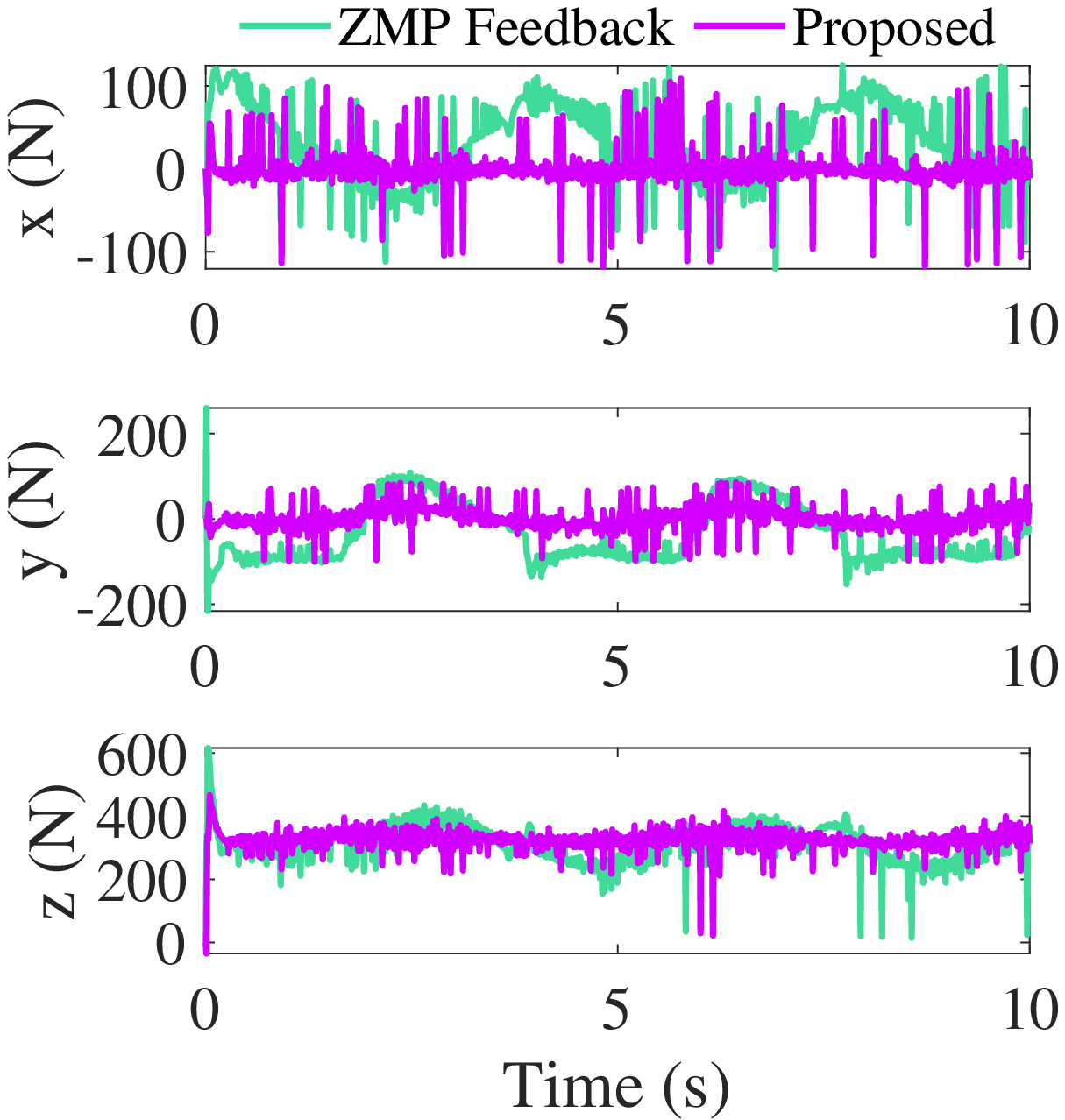}
    }
\subfigure[Moment on the left ankle]{
    \includegraphics[width=0.225\linewidth]{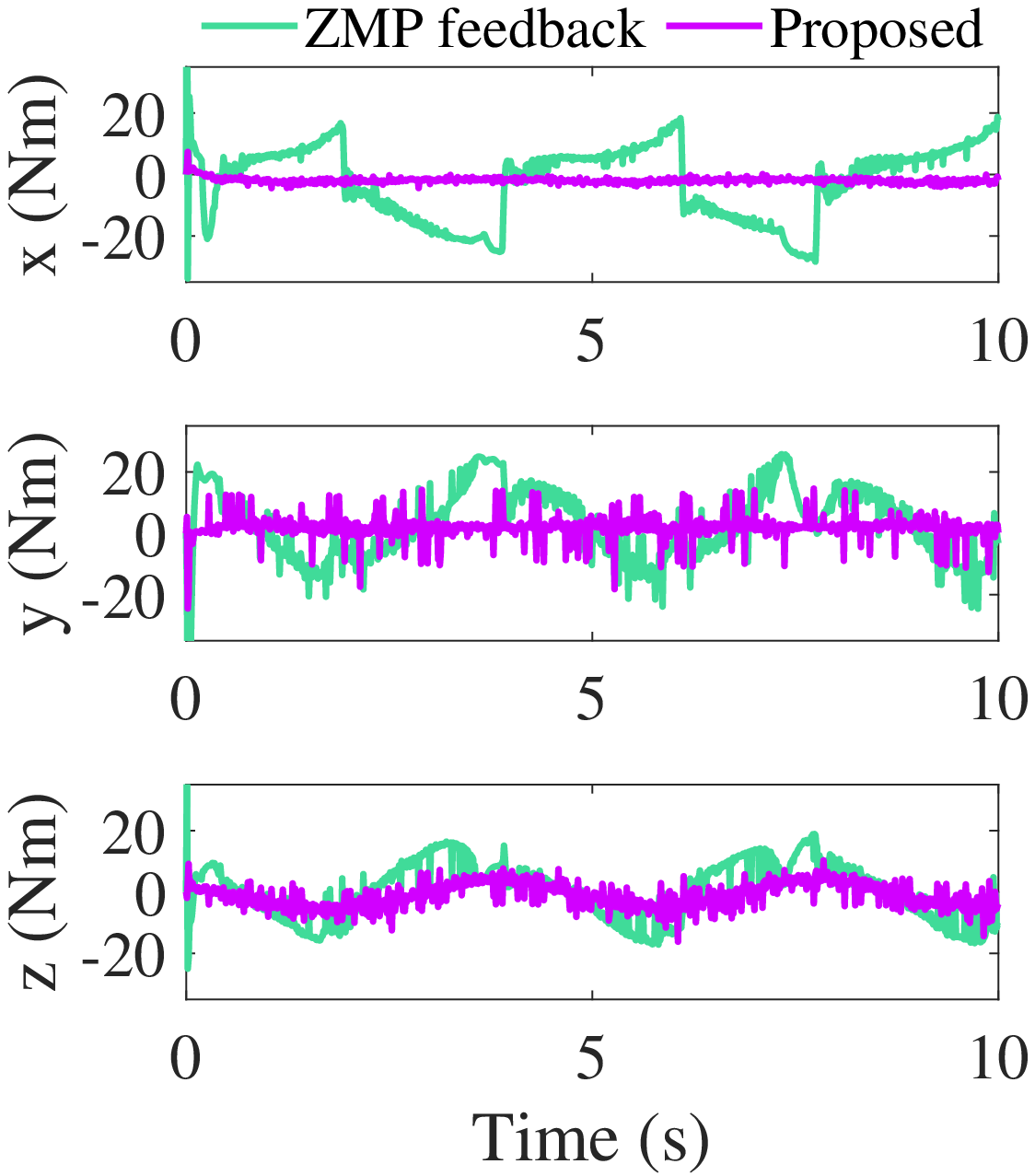}
    }
\caption{Floating-base trajectory and ankle wrench measurements corresponding to the scene shown in Fig.~\ref{fig:double}}
\label{fig: double_traj}
\end{figure*}

\begin{figure*}[htbp!]
    \centering
    \includegraphics[width=0.9\linewidth]{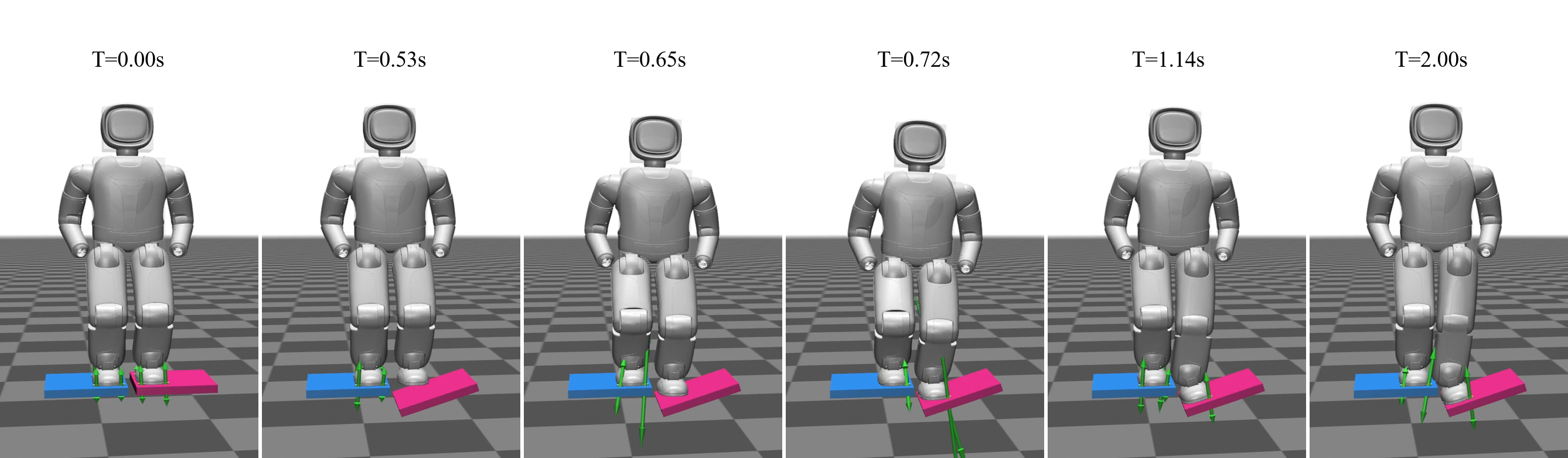}
    \caption{The platform under the left foot falls abruptly at the T=3.5s and its attitude is changed meanwhile.}
    \label{fig:down}
\end{figure*}

\begin{figure*}[htbp!]
\centering
\subfigure[Position of floating base]{
    \includegraphics[width=0.225\linewidth]{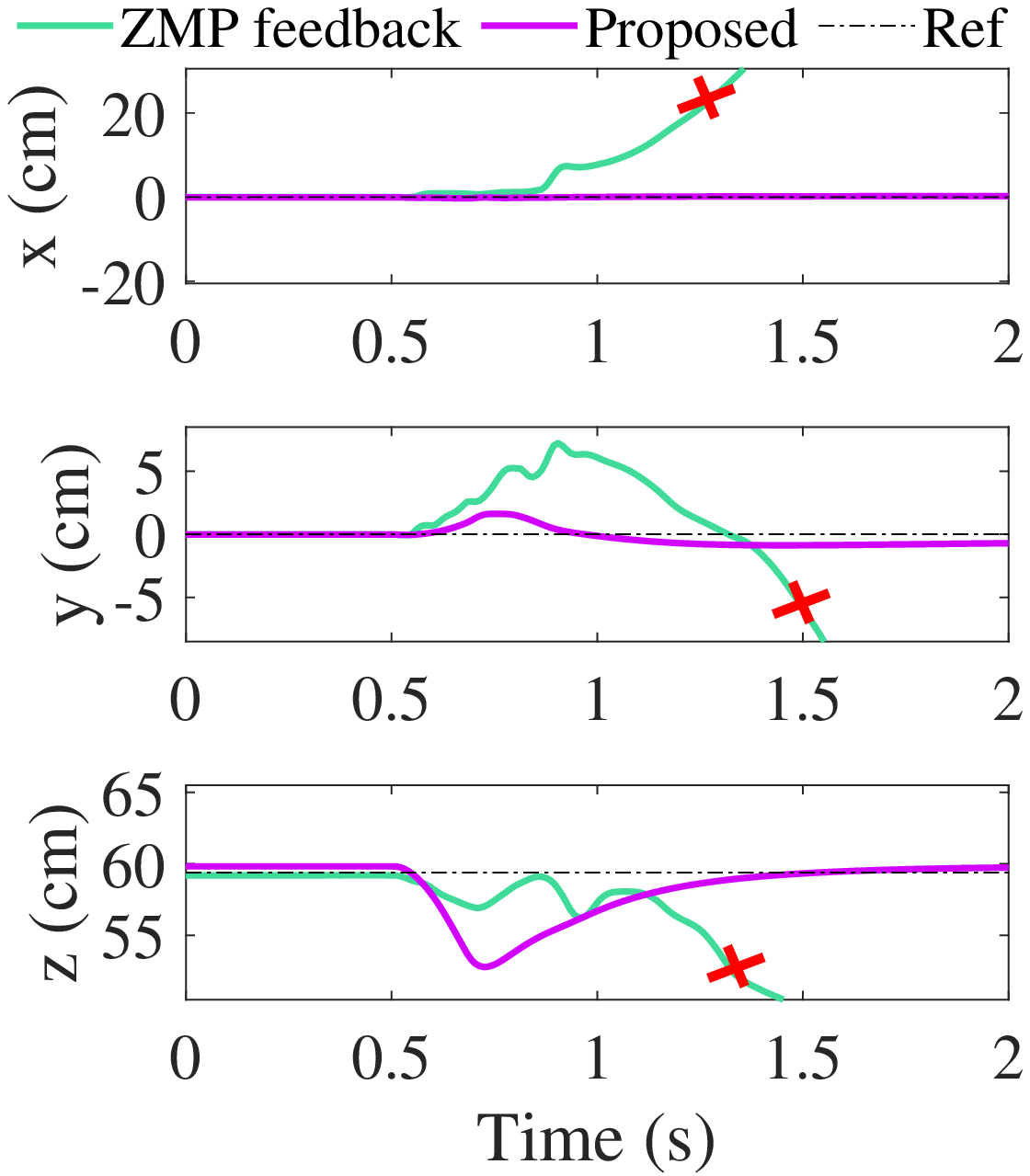}
    }
\subfigure[Rotation of floating base]{
    \includegraphics[width=0.225\linewidth]{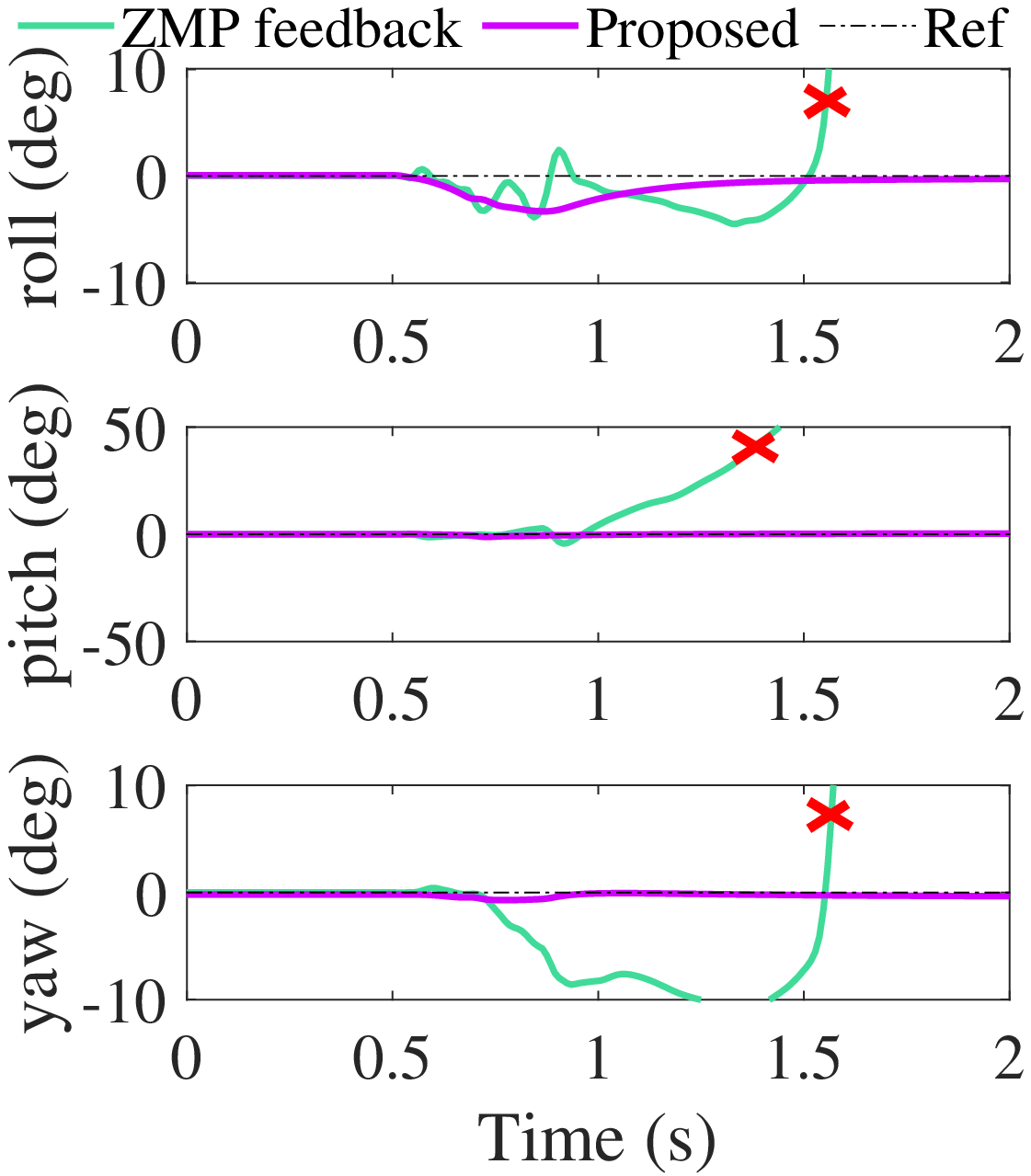}
    }
\subfigure[Force on the left ankle]{
    \includegraphics[width=0.235\linewidth]{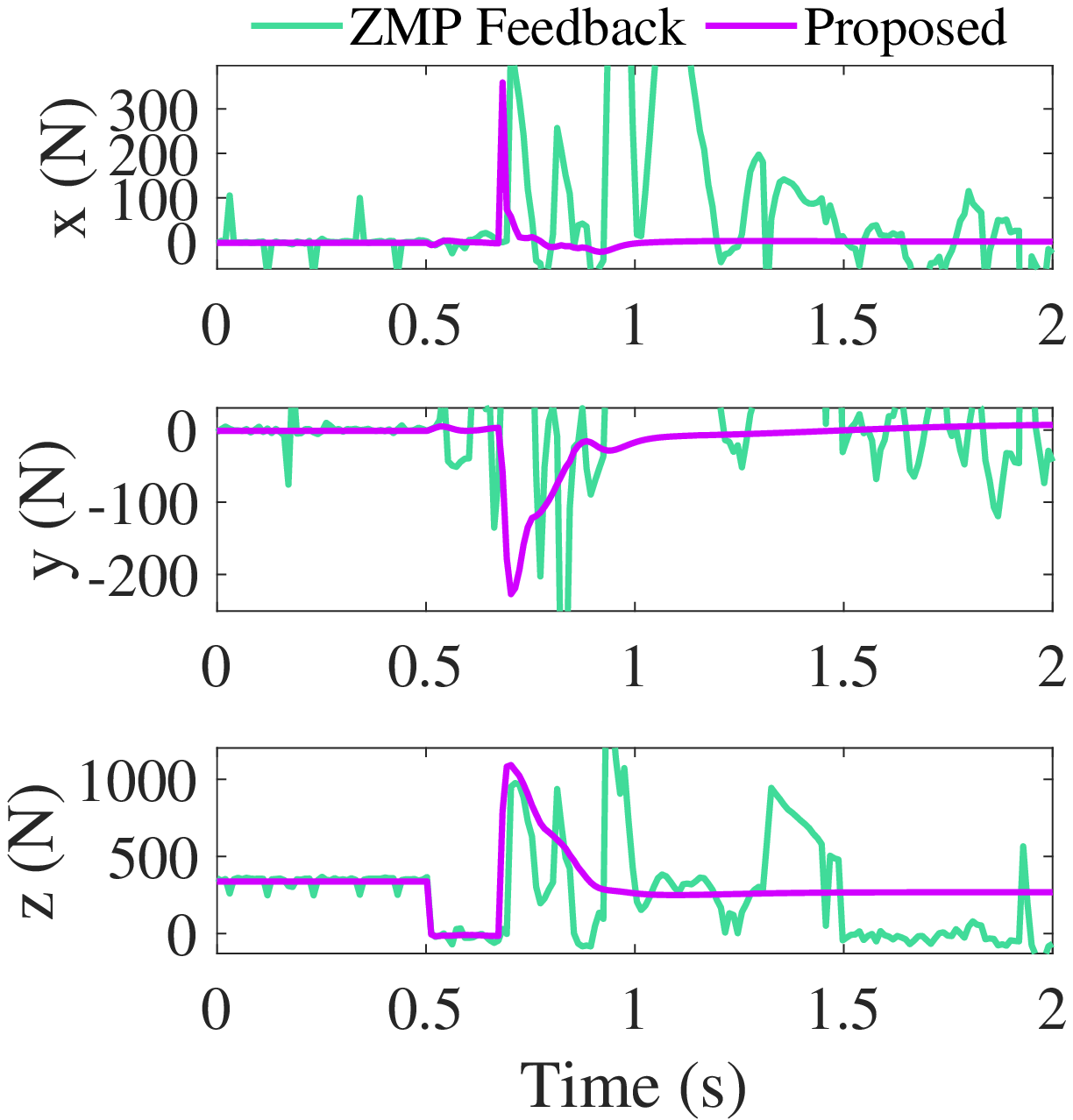}
    }
\subfigure[Moment on the left ankle]{
    \includegraphics[width=0.225\linewidth]{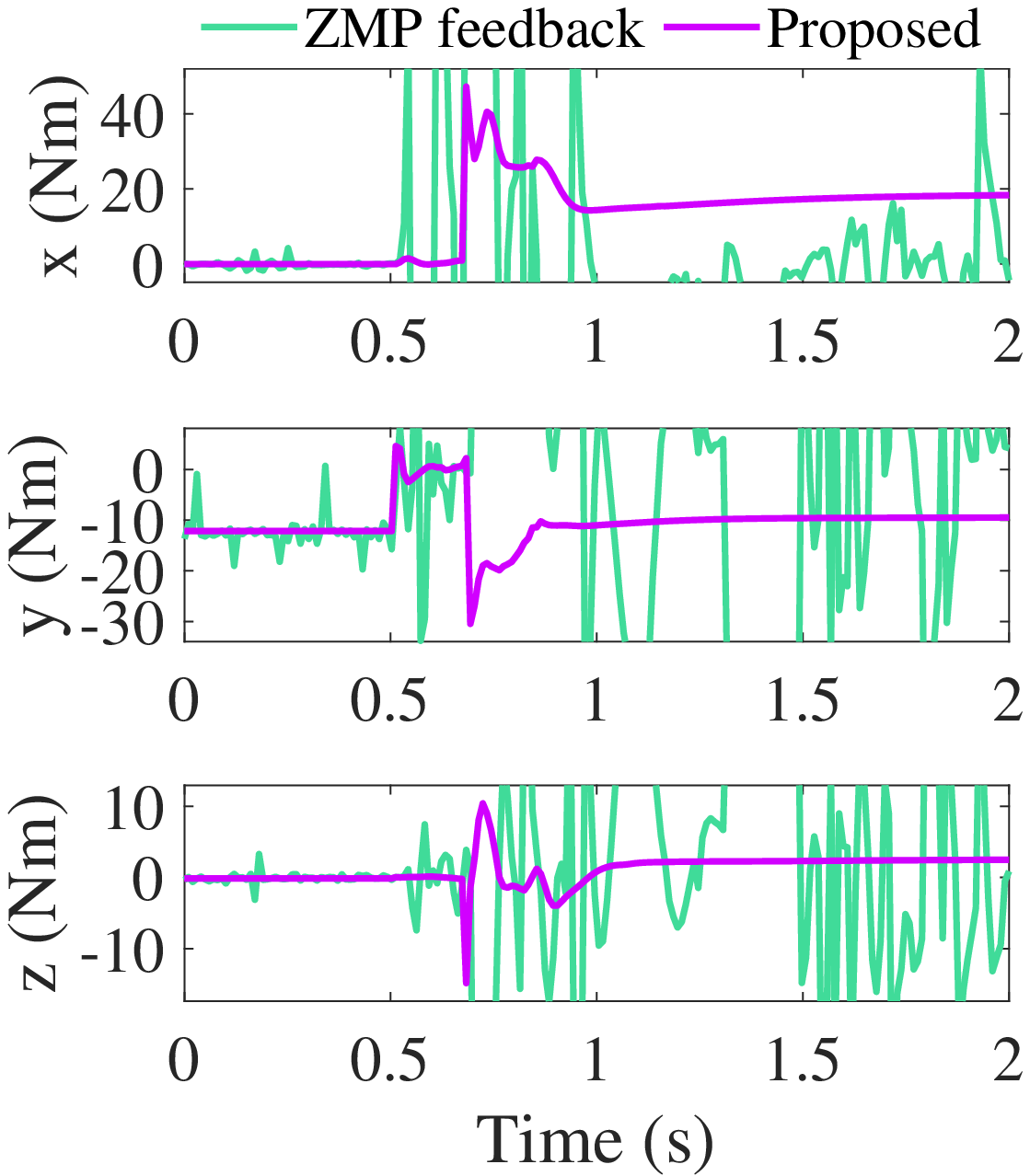}
    }
\caption{Floating-base trajectory and ankle wrench measurements corresponding to the scene shown in Fig.~\ref{fig:down}}
\label{fig: down_traj}
\end{figure*}

To test the effectiveness of the proposed stabilizer, we first consider a standard testing scenario in which the robot is placed on a moving platform whose height and orientation change continuously. When the height and the roll/pitch angle of the platform change, the contact force will change accordingly. To counteract this effect, the robot needs to adjust its joint inputs to achieve the desired contact force and to balance its CoM. 

Fig.~\ref{fig:heightAndRot} shows the snapshots of the simulation results where the robot balances itself on a continuously changing platform with the proposed stabilizer. In this situation, when the height and attitude are changed slowly, the baseline stabilizer with ZMP based feedback strategy achieves successful balancing as well, as illustrated in Fig.~\ref{fig: hr_traj}. From the figure, it is obvious that, as compared with the baseline approach, the proposed stabilizer achieves better performance with much smaller CoM variations and smaller contact forces. However, when the height and attitude change at a faster rate, the baseline stabilizer fails while the proposed stabilizer still achieves a stable balance.

To further demonstrate the strength of the proposed stabilizer, we test a slightly more challenging scenario, in which the terrain under each foot of the robot changes independently. In this case, effectiveness of the ZMP-based strategies heavily rely on the so-called ZMP distribution scheme using simplified model~\cite{5651082} , while the proposed stabilizer works without any additional tuning. Fig.~\ref{fig:double} shows the snapshots of the simulation with our proposed stabilizer. Similar to the previous case, when the terrain changes slowly, the ZMP-based stabilizers can still achieve balance. However, the tracking performance is worse than the proposed stabilizer, as shown in Fig.~\ref{fig: double_traj}. The proposed stabilizer remains capable of addressing scenarios with much higher terrain change rate as compared with the classical ones, where these results are available in the supplementary video.

For the considered scenarios where the terrain is changing continuously, the simulation validations indicate that the proposed force-feedback based whole-body stabilizer is capable of addressing much more complicated and challenging cases and achieves a better performance than the classical ZMP-based strategies in cases when both schemes are effective. 

\subsection{Scenario \uppercase\expandafter{\romannumeral2}: Reaction to Unexpected Discontinuous Terrain Change}
Now, we consider a fundamentally more challenging scenario where the terrain changes discontinuously. In particular, the scenario in which one of the support platform abruptly declines and changes in attitude. In this case, the contact between the support platform and the humanoid's foot breaks unexpectedly to the humanoid. 

Snapshots of the simulation with our proposed stabilizer is shown in Fig.~\ref{fig:down} and the associated CoM trajectories as well as contact force profiles are provided in Fig.~\ref{fig: down_traj}. Due to the break of original contact configuration and the non-coplanar contact situation after the terrain change, the classical ZMP-based stabilizer fails to provide a reasonable adjustment to the input commands to the humanoid to restore contact and balance. On the other hand, thanks to the utilization of six-dimensional contact force measurement and the coordination of whole-body motion, the proposed stabilizer successfully balances the robot and retains contact.



\subsection{Discussions}
The simulation experiments verify that taking into account of the full six-dimensional force measurement and the whole-body dynamical effects in designing stabilizers leads to a significant performance improvement as compared with conventional ZMP-based stabilizers. Moreover, the proposed stabilizer design strategy enables the humanoid to deal with more challenging scenarios that existing ZMP-based cannot handle.  Despite the impressive performance, the proposed stabilizer displays limitations in situations where the terrain structure changes too abruptly (larger than approximately $60\deg/s$ roll/pitch rate of change), which mainly attributes to the limited bandwidth of the position-controlled actuators. The reader is kindly referred to the supplementary video for additional simulation results showcase the comparisons with existing approaches under various settings and the limiting cases of the proposed stabilizers.

\section{Conclusion}
This paper concerns with stabilizer design for position-controlled humanoid robots. By capitalizing on the six-dimensional force measurement and accounting for the whole-body dynamics of the humanoid robot, we devise a novel force-feedback based whole-body stabilizer. The developed stabilizer involves task-space inverse dynamics problem and a task-space differential inverse kinematics problem that both can be efficiently solved via quadratic programming based techniques. Demonstrated by simulation experiments, the proposed stabilizer significantly improves tracking performance of position-controlled humanoid robots, and moreover enables the robots to tackle challenging contact scenarios. 

Potential extensions of this work include integration of the proposed stabilizer with high-level planners to achieve robust dynamic locomotion (e.g., walking and running) with position-controlled humanoid robots, and further validations of the proposed stabilizer on hardware platforms. 

\bibliographystyle{ieeetr}
\bibliography{mybibfile}

\end{document}